\documentclass[11pt,a4paper]{article}
\usepackage[hyperref]{eacl2021}
\usepackage{times}
\usepackage{latexsym}

\usepackage{microtype}

\aclfinalcopy % Uncomment this line for the final submission
\def\aclpaperid{610} %  Enter the acl Paper ID here

\usepackage{xcolor}
\usepackage{verbatimbox}
\usepackage[10pt]{moresize}
\usepackage{verbatim}
\usepackage{fancyvrb}
\usepackage{natbib}
\usepackage{graphicx}
\usepackage{bm}
\usepackage{booktabs}
\usepackage{multicol}
\usepackage{comment}
\usepackage{amsmath}
\usepackage{amsfonts}
\usepackage{amssymb}
\usepackage{xspace}
\usepackage{subfig}
\definecolor{amethyst}{rgb}{0.6, 0.4, 0.8}
\definecolor{darkorchid}{rgb}{0.6, 0.2, 0.8}
\definecolor{electricindigo}{rgb}{0.44, 0.0, 1.0}
\definecolor{persianblue}{rgb}{0.11, 0.22, 0.73}

\definecolor{myblue}{RGB}{57,73,171}
\definecolor{myamethyst}{RGB}{142, 3, 163}

\newcommand{\MF}{$\mathcal{M}\mathcal{F}_\beta$~}
\newcommand{\resmatch}{\textsc{Re\-Smatch}\xspace}

\DeclareSymbolFont{extraup}{U}{zavm}{m}{n}
\DeclareMathSymbol{\vardiamond}{\mathalpha}{extraup}{87}
\DeclareMathSymbol{\varheart}{\mathalpha}{extraup}{86}

\definecolor{airforceblue}{rgb}{0.36, 0.54, 0.66}
\definecolor{britishracinggreen}{rgb}{0.0, 0.26, 0.15}
\definecolor{byzantine}{rgb}{0.74, 0.2, 0.64}
\definecolor{cadmiumgreen}{rgb}{0.0, 0.42, 0.24}
\definecolor{lemonchiffon}{rgb}{1.0, 0.98, 0.8}
\definecolor{lavender(web)}{rgb}{0.9, 0.9, 0.98}
\definecolor{lightgreen}{rgb}{0.56, 0.93, 0.56}
\definecolor{lightcoral}{rgb}{0.94, 0.5, 0.5}
\definecolor{dollarbill}{rgb}{0.52, 0.73, 0.4}
\definecolor{etonblue}{rgb}{0.59, 0.78, 0.64}
\definecolor{lightsalmon}{rgb}{1.0, 0.63, 0.48}
\usepackage{amsmath}
\usepackage{amsthm}

\newtheorem*{pom}{Principle of meaning $\mathcal{M}$}
\newtheorem*{pof}{Principle of form $\mathcal{F}$}
\newtheorem*{prop}{Proposition}

\title{Towards a Decomposable Metric for Explainable Evaluation \\ of Text Generation from AMR}

\author{Juri Opitz \\
  Dept.\ of Computational Linguistics \\
  Heidelberg University \\
  69120 Heidelberg \\
  \texttt{opitz@cl.uni-heidelberg.de} \\\And
  Anette Frank \\
  Dept.\ of Computational Linguistics \\
  Heidelberg University \\
  69120 Heidelberg \\
  \texttt{frank@cl.uni-heidelberg.de} \\}

\date{}

\begin{document}

\maketitle

\begin{abstract}
    Systems that generate natural language text from abstract meaning representations such as AMR are typically evaluated using automatic surface matching metrics that compare the generated texts to reference texts from which the input meaning representations were constructed. We show that besides well-known issues from which such metrics suffer, an additional problem arises when applying these metrics for AMR-to-text evaluation, since an abstract meaning representation allows for numerous surface realizations. In this work we aim to alleviate these issues by proposing $\mathcal{M}\mathcal{F}_\beta$, a decomposable metric that builds on two pillars. The first is the \textbf{principle of meaning preservation $\mathcal{M}$}: it measures to what extent a given AMR can be reconstructed from the generated sentence using SOTA AMR parsers and  applying (fine-grained) AMR evaluation metrics to measure the distance between the original and the reconstructed AMR. The second pillar builds on a \textbf{principle of (grammatical) form $\mathcal{F}$} that measures the  linguistic quality of the generated text, which we implement using SOTA language models. In two extensive pilot studies we show that fulfillment of both principles offers benefits for AMR-to-text evaluation, including explainability of scores. Since \MF does not necessarily rely on gold AMRs,  
    it may extend to other text generation tasks.
\end{abstract}

\section{Introduction}\label{sec:intro}

\begin{figure}
    \centering
    \includegraphics[trim={0 0 0 0},clip,width=0.45\linewidth]{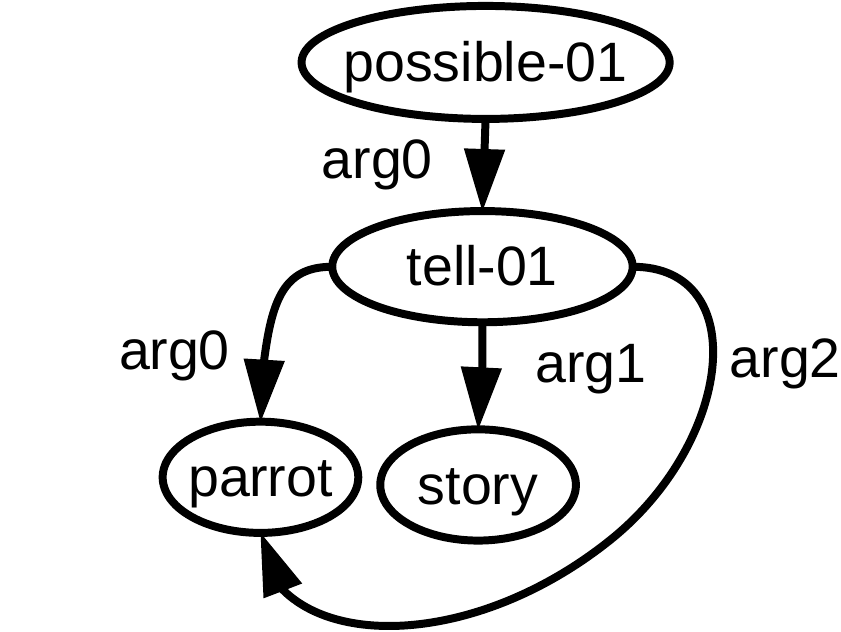}
    \caption{\textit{``Perhaps, the parrot is telling itself a story''}.
    }
    \label{fig:ex1}
\end{figure}

Abstract Meaning Representation (AMR, \citet{banarescu2013abstract}) aims at capturing the meaning of a sentence in a machine-readable graph format. AMR captures, i.a., word senses, semantic roles and coreference. The AMR in Fig.\ \ref{fig:ex1} represents the sentence \textit{Perhaps, the parrot is telling itself a story}. 
  In this graph, tell-01 links to a PropBank  \cite{palmer-etal-2005-proposition} frame, and arg$_n$ labels indicate participant roles: \textit{parrot} is both \textit{speaker} (arg0) and \textit{hearer} (arg2), \textit{story} is the \textit{utterance} (arg1). %Moreover, AMR can also model coreference (using re-entrancies, c.f., \textit{parrot} has two incoming edges) and speaker mode (in the example: \textit{interrogative}).% or polarity (via \textit{polarity} edges) and allows KB-linking (via \textit{wiki} edges). 

The task of AMR-to-text generation has recently garnered much attention %over the recent years 
\cite{song-etal-2017-amr,song-etal-2018-graph,konstas-etal-2017-neural,DBLP:conf/aaai/CaiL20,ribeiro-etal-2019-enhancing}. The output of AMR-to-text systems  %generated from AMRs are
is
typically evaluated against the sentence from which the AMR was created, using standard surface string matching metrics such as \textsc{Bleu} \cite{papineni-etal-2002-bleu} or \textsc{chrF(++)} \cite{stanojevic-etal-2015-results,popovic-2015-chrf, popovic-2016-chrf, popov-2017-word}, as is standard in many
%employed in general 
NLG tasks.  %Yet,
%However, 
These metrics suffer from several issues, for example, they penalize paraphrases, are highly sensitive to outliers \cite{mathur-etal-2020-tangled}, and lack interpretability \cite{nemasai2020survey}. 

\begin{figure}
    \centering
    \includegraphics[width=1.0\linewidth]{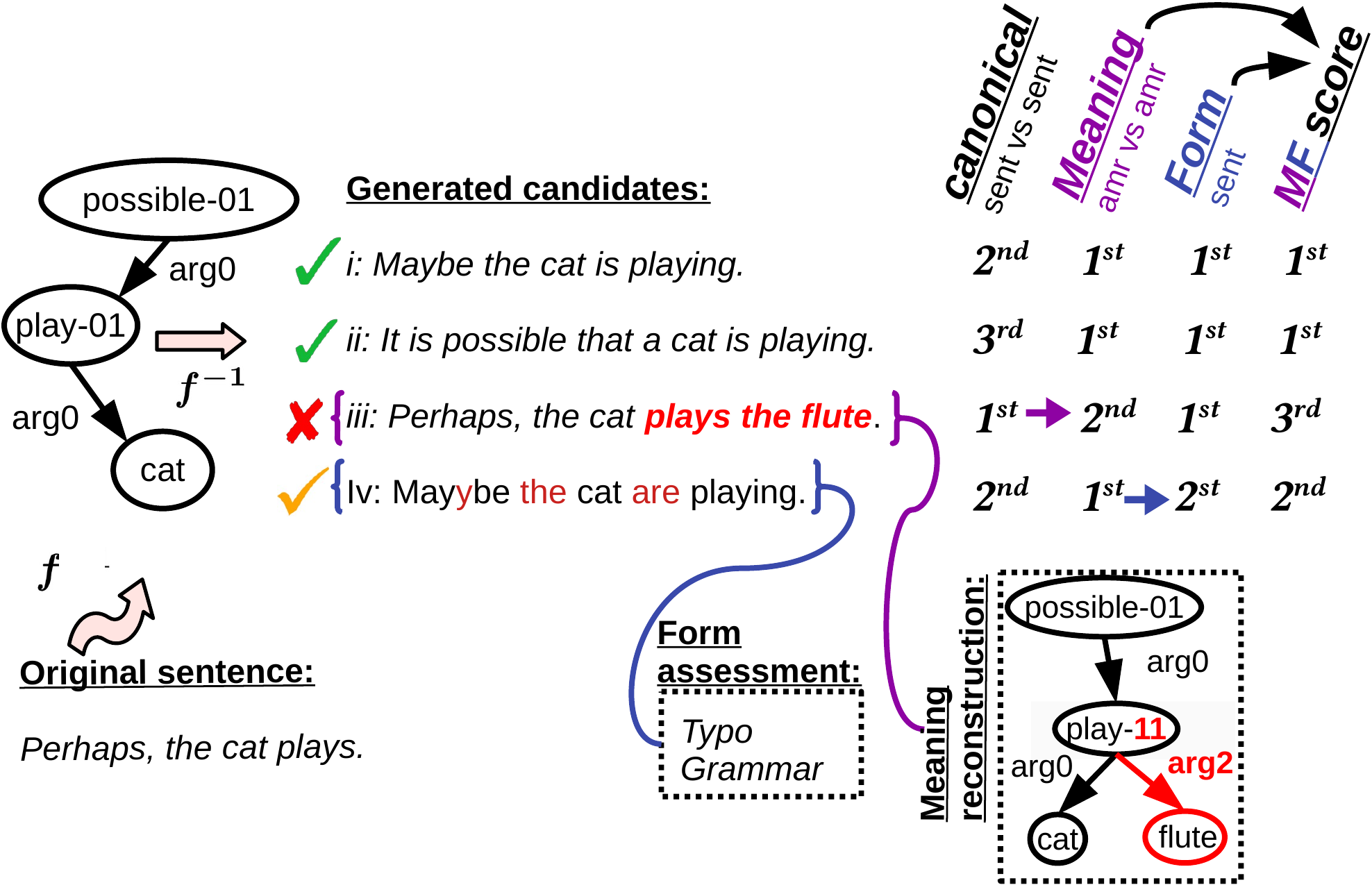}
    \caption{The \underline{\textbf{\textit{Canonical}}} evaluation matches n-grams from the sentences and assigns inappropriate ranks. Our metric \MF fuses \textcolor{myamethyst}{\underline{\textbf{\textit{Meaning}}}} and \textcolor{myblue}{\underline{\textbf{\textit{Form}}}} assessment and better reflects the ranking of the generations.}
    \label{fig:problem1}
\end{figure}

%We find that some 
Some of these issues 
%become aggravated 
get compounded when evaluating AMR-to-text. The core of the 
problem %The root cause 
%lies in the fact 
is that there are many
%manifold 
ways to realize a sentence from a meaning representation. 
%For example, 
Fig.\ \ref{fig:problem1} shows four candidate sentences (\textbf{i}-\textbf{iv}) for a given AMR (left). 
%In this case, 
One system generates \textbf{(i)}: \textit{Maybe the cat is playing.} while another generates \textbf{(iii)}: \textit{Perhaps, the cat plays the flute}. Clearly, \textbf{(i)} 
%better 
captures the meaning 
%contained in
of the gold graph better than
%(left) compared to 
\textbf{(iii)}, which contains `hallucinated' content -- 
a well-known
%a severe 
issue in neural generation
%models
 \cite{logan-etal-2019-baracks,wang2020exposurehallu}. 
%Now, 
Yet, when using a canonical metric such as \textsc{Bleu}
%\footnote{with NIST sentence smoothing \cite{chen-cherry-2014-systematic}.} 
to evaluate 
%the generated 
sentences
%for evaluating the \underline{original} against the \underline{generated} sentence, the system that. 
% Yet, when comparing 
\textbf{(i)} and \textbf{(iii)} against the 
%original sentence, 
reference, the system that 
%sticks true to the meaning of the AMR is penalized, while the system that
produces 
%the hallucinating sentence
hallucinations (\textbf{iii}) is greatly rewarded (54 \textsc{Bleu} points) to the disadvantage of systems that yield meaning preserving sentences (\textbf{i}) (18 points) and  (\textbf{ii}) (5 points). 

This work aims at a (better) metric that \textbf{measures meaning preservation} of the generated output towards the MR given as input,
%; we do this 
by (re-)con\-structing an AMR from the generated sentence and comparing it to the input AMR. In Fig.\ \ref{fig:problem1}, \underline{Reconstruction} is the result of parsing \textbf{(iii)}. The reconstructed AMR exposes several
%the 
meaning deviations 
%as 
(marked in red):
%in the sense that it deviates from the original MR (marked in red). Specifically, 
%\textbf{iii} 
%misrepresents the
it contains an alternate sense of \textit{play} 
%(01 vs.\ 11) 
and contains an additional semantic role arg2 with filler
%plus its content 
\textit{flute}. By contrast, when converting sentences \textbf{(i)}, \textbf{(ii)}, or \textbf{(iv)} to AMRs, we obtain flawless reconstructions. %To fuse these outcomes into scores, we may apply well-defined graph-matching metrics (AMR vs.\ AMR). 
We will measure preservation of \textcolor{myamethyst}{\underline{\textbf{\textit{Meaning}}}}  using well-defined graph matching metrics.

Figure \ref{fig:problem1} also illustrates that 
%Then, we also see that 
assessing meaning preservation is not sufficient to rate
%may not be enough to be fully capable of rating
the quality of generations: (\textbf{iv}) captures the meaning of the AMR well -- but its form is flawed: it suffers from wrong verb inflection, a common issue %especially 
in low-resource text generation settings \cite{koponen2019product}. 

In order to rate both meaning and form of a generated sentence, 
%To resolve this issue,
we 
%merge 
combine the score for meaning 
reconstruction with a score called \textcolor{myblue}{\underline{\textbf{\textit{Form}}}} that 
%allows us to 
\textbf{judges the sentence's grammaticality and fluency}.
%, short: \textcolor{myblue}{\underline{\textbf{\textit{Form}}}}. 
 By these moves, we obtain a more suitable and explainable ranking with a combined \textcolor{myamethyst}{\underline{\textbf{\textit{M}}}}\textcolor{myblue}{\underline{\textbf{\textit{F}}}}\underline{ score}.\footnote{See Fig.\ \ref{fig:problem1}: 1$^{st}$/2$^{nd}$ rank: \textbf{i}; 3$^{rd}$ rank: \textbf{iv}; 4$^{th}$ rank: \textbf{iii}.} By clearly distinguishing between \textcolor{myamethyst}{\underline{\textbf{\textit{Meaning}}}} and  \textcolor{myblue}{\underline{\textbf{\textit{Form}}}}, our  \textcolor{myamethyst}{\underline{\textbf{\textit{M}}}}\textcolor{myblue}{\underline{\textbf{\textit{F}}}}\underline{ score} (henceforth denoted by $\mathcal{M}\mathcal{F}_\beta$) also aligns well with recent calls to achieve a clearer separation of these aspects in NLU \cite{bender-koller-2020-climbing}. 
 
 Generally, our contributions are as follows:

\paragraph{(1)} We propose two linguistically motivated principles that aim at a sound evaluation of AMR-to-text systems: the \textbf{principle of meaning preservation} and the \textbf{principle of (grammatical) form}.

\paragraph{(2)} From these 
%complementary 
principles we derive and implement a (novel) \textbf{\MF score for AMR-to-text generation}\footnote{We make code available at \url{https://github.com/Heidelberg-NLP/MFscore}.} which is composed 
%its score based on 
of individual 
%measurements of 
metrics for meaning and form aspects. \MF allows users to modulate these two views on generation quality 
%with respect to
to vary their impact on the final metric score.

\paragraph{(3)} We conduct two major pilot studies involving (English) text generations from a range of competitive AMR-to-text systems and human annotations. 
%In 
First we study the potential practical benefits of \MF when evaluating systems, such as its prospects to offer interpretability of scores and finer-grained system analyses. The second study probes potential weak spots of \MF, e.g., its dependence on a strong AMR parser.  

\paragraph{} We consider \MF as it stands
%can be readily applied 
as a suitable metric to enhance interpretability of generation scores.

\if false
\begin{itemize}
\itemsep0em 
  %  \item We linguistically motivate two principles that aim at ensuring researchers a sound evaluation of generation from meaning representation.
    \item We propose two linguistically motivated principles that aim at a sound evaluation of AMR-to-text systems: (i) the principle of meaning preservation and (ii) the principle of (grammatical) form.
    \item From these complementary principles we derive and implement a (novel) \textbf{\MF score for AMR-to-text generation} which composes its score based on individual measurements of meaning and form aspects. \MF allows users to modulate these two views on generation quality with respect to their impact on the final metric score.\footnote{We will release all code and data.}%, allowing for, e.g., custom gauging of AMR-to-text systems
    \item We conduct two major pilot studies involving a range of competitive AMR-to-text generation systems and human annotations. In the first study, we investigate the potential practical benefits of \MF when assessing systems, such as its prospects to offer interpretability of metric scores and finer-grained system analyses. In the second study, we assess potential weak spots of \MF, for example, its dependence on a strong AMR parser.  
\end{itemize}
\fi

\section{Fusing meaning and form into \MF}\label{sec:formalproblem}

While current NLG metrics lack \textit{interpretability} and mainly focus on the form of generated text \cite{nemasai2020survey}, in this work we emphasize the \textit{meaning aspect} in NLG evaluation, which is most clearly dissociated from form %in systems that perform NLG
when generating text
from structured inputs such as AMR. At the same time, form and wording of the  generated text cannot be ignored, as we want such systems to produce \textit{natural
%,  fluent 
and  well-formed sentences}.
%Then, with this knowledge, 
Equipped with this two-fold objective,
%this background, 
we start building our \MF score which aims at a \textit{ balanced combination} of both quality aspects: \textbf{meaning and form}.

\subsection{From principles to \MF} 
%To alleviate the issue described above, 
In a first step we introduce our

\begin{pom}
Generated sentences should allow loss-less AMR reconstruction. 
\end{pom}

This principle expresses a key expectation 
%that we formulate 
for
%pose to 
a system that generates NL sentences
%generates text 
from abstract meaning representations. Namely, the generated sentence should reflect the meaning of the AMR. So, in order to 
%really 
assess whether a generated sentence $s' = f^{-1}(m)$ 
%and 
is
%both 
a valid generation for 
%outputs from 
the input AMR $m$, rather than matching $s'$ against a reference sentence $s$,
%input structure, 
we 
%need to
 perform this assessment 
\textbf{in the abstract MR domain},
%This can be achieved 
by applying an inverse system $f$ that
\textit{parses} the generated text back to an AMR $m' = f(s') = f(f^{-1}(m))$.
%generates AMR from text (a parser). 
I.e., we desire a 
$metric: \mathcal{D} \times \mathcal{D} \rightarrow [0,1]$ that satisfies: 
%the condition: 
$s \equiv s' \iff m = m' \iff metric(s,s') = 1$.
Two texts are equivalent iff their meaning abstractions %lead to 
denote the same meaning.
%abstract meaning construction. 
In case $f(s')$ yields an AMR $m' \neq m$, we can still determine the degree to which $s'$ preserves the meaning of AMR $m$ by measuring the distance between $m$ and $m'$ by standard AMR metrics, e.g.,  $Smatch(m,m')$. 

Note that computing $Smatch(m,m')$ does not depend on a reference sentence, because the comparison is conducted purely in the abstract domain. This is mathematically more appealing for the evaluation of AMR-to-text, since it solves the problem that one abstract representation may result in various (valid) surface realizations  (cf.\ Appendix \ref{sec:appendix-proof}). Finally, we also do not necessarily need to rely on
%depend on 
a gold graph $m$,
%either, 
but can instead set $m = f(s)$, i.e., the parse of the reference sentence. This means that future application of $\mathcal{M}$ to other kinds of text generation tasks is straightforward.

However, 
%we may note that 
the principle $\mathcal{M}$ alone is not sufficient: we also expect the system to generate grammatically well-formed and fluent text. For example, $s'$: 
%following system output: 
\textit{Possibly, it(self) tells parrot a story.}  contains relevant content expressed in the AMR of Fig.\ \ref{fig:ex1}, but it is neither grammatically well-formed, nor a natural and fluent sentence.
This leads us to our

\begin{pof}
Generated sentences should be syntactically well-formed, natural and fluent.
%have acceptable surface structures. 
\end{pof}

In the style of the well-established $F_\beta$ score \cite{DBLP:books/bu/Rijsbergen79}, we fuse these two principles into the $\mathcal{M}\mathcal{F}_\beta$ score:
\begin{equation}
\label{eq:metricmain}
\hspace*{-2mm}    \mathcal{M}\mathcal{F}_\beta = (1+\beta^2) \frac{Meaning \times Form}{(\beta^2 \times Meaning) + Form}
\end{equation}

%Here, 
Here, $Form$ and $Meaning$ are expressed as ratios that will be more closely described in the following subsection. $\beta$ allows users to gauge the evaluation towards $Form$ or $Meaning$, depending on %accounting for 
specific application scenarios.
%We anticipate that most 
Users 
%will
may prefer the harmonic mean ($\beta$ = 1) or may
%will 
give $Meaning$ double weight %compared to 
compared to
$Form$ (e.g., 
%by setting 
$\beta$ = .5).\footnote{Generally, $Form$ receives $\beta$ times as much importance compared with $Meaning$.}
%In general, we anticipate that most researchers may want to put emphasis on $Meaning$ while also not completely neglecting the $Form$. This setting can be expressed well with $\beta=0.5$. 
%However, i
In our experiments we
%will 
%also 
consider extreme decompositions into $Meaning$-$only$ ($\beta \rightarrow 0$) or $Form$-$only$ ($\beta \rightarrow\infty$).
%and their harmonic mean ($\beta = 1$). 

\subsection{Parameterizing meaning} \label{sec:para_meaning}
We 
%propose to 
measure $\mathcal{M}$ or $Meaning$ (Meaning Preservation)
%$meaning$ 
with a score range in
%\in 
$[0,1]$ by reconstructing the AMR with a SOTA
%state-of-the-art 
parser and computing the relative graph overlap of the reconstruction and the source AMR using graph matching. We call this \resmatch.
%(short: \resmatch). 
%I.e., g
Given a generated sentence $s'$ and source AMR $m$, we match $parse(s')$ against $m$ 
%and compute $Meaning =
%$ as 
by computing $
amrMetric(parse(s'),m)$. This means that we have to decide upon $parse$ and $amrMetric$. %Next, w
We propose two potential settings.

\paragraph{AMR reconstruction} To reconstruct the AMR with $parse$, we %will 
use the latest state-of-the-art AMR parser
%be using %propose to use 
%the parser 
by \citet{cai2020amr}. With 80.3 Smatch F1,
%\footnote{measured on the standard benchmarking corpus}
this parser is almost on-par with human agreement (estimated  at 0.71--0.83 Smatch F1 in \citet{banarescu2013abstract}). We henceforth call it GSII.

\paragraph{Assessing $\mathcal{M}$ 
%for reconstruction 
with AMR metrics}

To obtain a score for  $\mathcal{M}$ we propose to use  S$^2$match 
\cite{TACL2205} -- a variant of Smatch \cite{cai-knight-2013-smatch} that performs 
a \textit{graded match for concept nodes}. This offers the potential to compensate for noise 
in automatically generated text or minor lexical deviations from the original sentence.

\paragraph{Discussion} 
%All in all, 
Comparing to references
%sentences 
%in the domain of 
by matching their meaning graphs has the prospect of offering interpretability and explanations, by detecting redundant or missing meaning components in the generations.
%generated sentences. 
In our 
%experimental study,
studies, we will see that this assessment can be conducted 
%either 
by computing 
%\textit{one meaning graph overlap score} 
a \textit{single graph overlap score}
(e.g., 
%via 
S$^2$match F1), or along \textit{multiple dimensions of meaning}, such as SRL, coreference or WSD \citep{damonte-etal-2017-incremental}. Generally, \MF gives
researchers 
%with 
%much
%a lot of 
the flexibility
%as to 
of choosing a 
%which 
$parse$r or $amrMetric$ to their liking. 
%they %may prefer. 
%For our work, 
In this work, we choose
%our $parser$, we aimed at 
the %possibly 
best current $parse$r
%best one
that achieves high IAA with humans. 
Yet,
we would also like to know whether the $parser$ is vulnerable to specific peculiarities of generated sentences, or how using another parser affects the scores. 
%We would also like to gain knowledge about the outcome of using another parsing system. 
We will
%Therefore, we will 
investigate these issues more closely in \S \ref{subsec:parsequality}.  %\footnote{One may also fall back on graph metrics for other meaning representations, e.g., the evaluation of DRS \cite{kamp1981theory,kamp2013discourse} is conducted similarly to Smatch \cite{van-noord-etal-2018-evaluating} but suffers from more complexity due to larger graphs. \citet{liu-etal-2020-dscorer} therefore develop a more efficient metric that circumvents the costly alignment.} 

\subsection{Parameterizing form with LMs} \label{sec:para_form}
Assessing 
%the (related) aspects of 
sentence grammaticality and fluency is not an easy task \cite{heilman-etal-2014-predicting,katinskaia-ivanova-2019-multiple}. Recently, \citet{ 
lau2020furiously, zhu2020gruen} show
%, who show 
that probability estimates based on language models can be used as an indicator for measuring complex notions of form and for measuring acceptability in context. 
%Here, we 
For our \MF score we desire an interpretable ratio as input, which we base on 
%the 
LM predictions as follows.

\paragraph{Binary form assessment}

 Given a specific candidate generation $s'$, we use a binary variable to assess whether $s'$
 %the generation 
 is of satisfactory form. For this, we first calculate the mean token probability:\footnote{We use the mean (instead of the product) because \citet{bryant-briscoe-2018-language} find that basing decisions on the mean works well in practice when assessing possible corrections of grammatical errors.}
 
\begin{equation}
    mtp(\cdot)= \frac{1}{n}\sum_{j=1}^{n} P(tok_{j} | ctx_j),
\end{equation}

where $ctx_j$ is different for uni-directional LMs ($ctx_j = tok_{1...j-1}$) and bi-directional LMs ($ctx_j = tok_{1...j-1,j+1...n}$). We compute 
 $mtp$
%this score both 
for the generated sentence $s'$
% $mtp(s')$ 
and 
%for 
the reference $s$
 %source 
 %sentence as reference 
 %$mtp(s)$, calculating 
 and calculate a 
 %score of 
 preference score $prefScore= \frac{mtp(s')}{mtp(s')+mtp(s)}$. The 
decision of whether the $Form$ of 
 %decision on whether the 
 %shallow form of the 
 a generated sentence $s'$ is acceptable is 
 then calculated as 
\[
    accept= 
\begin{cases}
    1 ,& \text{if } prefScore \geq 0.5 - tol\\
    0,              & \text{otherwise},
\end{cases}
\label{eq:tol}
\]
where $tol$ is a tolerance parameter. Less formally, a sentence is considered to have an acceptable surface form in relation to its  
%as structurally acceptable 
%with respect to a 
reference if its form is estimated to be
%as being 
at least as good as the reference minus a tolerance, which we fix at 0.05.
%Finally, is m
I.e., the corpus-level $Form$ score reflects the ratio of generated sentences that are of acceptable form.\footnote{
%This means that 
I.e., the $Form$ score for a single sentence with $accept\geq 0.5 - tol$ equals 1.0. 
If a precise assessment for a single sentence is needed, we can fall back on $prefScore$ (+/- tol).}

\paragraph{Predictor selection}  We consider GPT-2 \cite{radford2019language}, distil GPT-2 \cite{sanh2019distilbert}, BERT \cite{devlin-etal-2019-bert}  and RoBERTa \cite{roberta} as a basis for assessing $Form$. We conduct experiments on WebNLG 
%data of 
\citep{gardent-etal-2017-webnlg,shimorina2017webnlg}, which contains human fluency and grammaticality judgements for machine-generated sentences. We find that GPT-2 performs best:
%Based on the results, 
%since we find that 
it 
discriminates sentences of poor and perfect fluency and grammaticality with an F1 score of approximately 0.8, and 
%it 
shows marginally better performance compared to the other LMs (see 
Appendix \ref{sec:appendix-webnlgexp} for the experiment details).
%gives details for this preliminary experiment.
We thus select GPT-2 as our LM
%basis 
for 
$Form$ assessment. 

\paragraph{Discussion} While  the reconstruction 
of meaning  
does not depend on the reference sentence, we do make use of 
it, in $prefScore$,
%the reference sentence 
for better assessment of $Form$. One reason 
%for this 
is that when assessing the form of 
a sentence $s'$ that contains 
rare words,
%that are unlikely to occur, 
the `raw' $mtp(s')$ 
%of the generation 
may be too pessimistic and may not well relate to the quality of the form. Generally, the $mtp$ (or any LM probability) itself is not well interpretable and hardly allows comparison to the $mtp$ of other sentences 
%(especially if they, 
(e.g., if they are about a different topic). 
%Therefore, 
%In such cases, the comparison to the reference enables us to gain a reliable score.
%only in consideration of a reference of which we assume it is grammatically well-formed and expresses similar content, we can say whether such a rare sentence is of acceptable form or not. 
However, by 
%By laxly/point-wise 
relating the $mtp$ of the generated sentence to the $mtp$ of a (same-topic) reference, we gain three advantages: first, we do not, a-priori, penalize generations that contain rare words. Second, we obtain an interpretable corpus-level ratio (rate of sentences that are of acceptable form). This is important, since sound \MF calculation ideally requires two interpretable ratios as input. Third, by avoiding any
%any
string matching, we still
keep 
%do not risk intermingling 
form and meaning aspects clearly distinct.

\subsection{Goals of our pilot studies}

%In \S 3.1 above  we outlined some theoretical advantages of \MF. 
Our main aim 
%with the proposed \MF metric for AMR-to-text generation 
is to establish, with the proposed \MF score for AMR-to-text generation,
%aims at 
%is aimed at offering 
 \textbf{i)} a \textbf{
balanced} 
%and justified 
and \textbf{interpretable assessment} of 
generated text  
%ed sentences 
according to \textbf{\textit{Meaning} and \textit{Form}}. Yet, as detailed in \S \ref{sec:para_meaning} and \S \ref{sec:para_form},  both components
%they 
depend on a number of \textbf{ii)} \textbf{hyperparameters},
%However, the computation of \MF is complex and depends on several hyper-parameters, 
such as the parser applied for $Meaning$ reconstruction, or the LM used %LMs 
for $Form$ assessment.
%Furthermore, 
%since 
These parameters may also be subject to change over time.
%(e.g., LM capacity or AMR parse performance). 
%an important task is 
It is thus important to assess the effects of such factors on metric scores and system rankings. 
%To investigate 
%shed more light onto 
%these questions, we conduct \textbf{two pilot studies}.
We investigate both aspects of \MF in \textbf{two pilot studies}.

In \textbf{the first 
%pilot 
study}, in \S 3, we 
aim
%want
to 
assess the prospects of \MF when ranking
SOTA
%state-of-the-art 
systems. 
%In particular,
We will see that \MF can explain system performance differences by disentangling $Form$ and $Meaning$, an 
%aspect 
asset that 
no 
%none of the 
other metric %provides. 
can offer.
%Here, we want to examine explanations for ranking differences on the sentence and the system level.

%In 
\textbf{The second 
%pilot
study}, in \S 4,
%, we 
investigates 
%potential drawbacks 
the impact of \MF's
%, namely its 
dependence on a parser and a LM. %Therefore,  
We i) investigate the effects of  using different
%another 
parsers,
%and
ii) 
%we 
assess the potential suitability of \MF for other text generation tasks, by ablating the human gold graph from the evaluation and using \MF to evaluate generated text vs.\ reference text, and %Finally, 
iii) %We 
validate the LM's binary predictions for
%by
$Form$ in 
%an
%small 
a manual annotation study.

\section{Study I: 
%Assessing %potential %advantages of \MF: 
Assessing interpretability}

\begin{table*}[htp!]
\centering
\scalebox{0.77}{
    \begin{tabular}{l|r|rrrr|rrrr|r|r}
    \toprule
    & & & & & &\multicolumn{3}{c}{$Meaning$} & $Form$ & $\mathcal{M}\mathcal{F}_1$ & $\mathcal{M}\mathcal{F}_{0.5}$\\
   & abbrev.\ &\textsc{Bleu}& \textsc{Meteor} & chrF++ & BERTsc.\ &\multicolumn{3}{c}{\resmatch} & - & - & -\\
     & &&&&F1&P&R&F1 & \%acc.\ & Eq.\ \ref{eq:metricmain} &  Eq.\ \ref{eq:metricmain}\\
     \midrule
     $apprUB$ & - & - & - & - & - & 83.1 & 80.1 & 81.5 & 100 & 89.8 & 84.6\\ 
     \midrule
      \citet{ribeiro-etal-2019-enhancing} & R'19 & 27.9$_{\rm{(5)}}$ & 33.2$_{\rm{(7)}}$ & 58.7$_{\rm{(6)}}$& 92.7$_{\rm{(4)}}$ & 76.5 & 67.7 & 71.9$_{\rm{(6)}}$  & 51.6$_{\rm{(5)}}$ & 60.1$_{\rm{(5)}}$ & 66.6$_{\rm{(5)}}$\\
       \citet{guo-etal-2019-densely}& G'19  & 27.6$_{\rm{(6)}}$ &33.7$_{\rm{(6)}}$ &57.3$_{\rm{(7)}}$& 92.4$_{\rm{(7)}}$& 78.2 & 70.0 & 73.9$_{\rm{(3)}}$  & 47.1$_{\rm{(7)}}$ & 57.5$_{\rm{(7)}}$ &66.3$_{\rm{(6)}}$\\

       \citet{TACL1805} & Wb'20 & 27.3$_{\rm{(7)}}$ & 34.1$_{\rm{(5)}}$ & 59.3$_{\rm{(5)}}$ & 92.6$_{\rm{(6)}}$& 79.6& 65.0& 71.5$_{\rm{(7)}}$ & 49.5$_{\rm{(6)}}$ & 58.5$_{\rm{(6)}}$ & 65.7$_{\rm{(7)}}$\\
       \citet{DBLP:conf/aaai/CaiL20} & C'20 & 29.8$_{\rm{(4)}}$  & 35.1$_{\rm{(4)}}$ & 59.4$_{\rm{(4)}}$ & 92.7$_{\rm{(4)}}$& 78.1 & 69.2 & 73.4$_{\rm{(5)}}$ & 51.9$_{\rm{(4)}}$ & 60.3$_{\rm{(4)}}$ &67.0$_{\rm{(4)}}$ \\
       \citet{mager-etal-2020-gpt-too}-M &Mb'20 & 33.0$_{\rm{(2)}}$ & 37.3$_{\rm{(2)}}$ & 63.1$_{\rm{(3)}}$ & 93.9$_{\rm{(2)}}$&79.4 & 68.7 & 73.7$_{\rm{(4)}}$  & \textbf{74.0}$_{\rm{(1)}}$& \textbf{73.9}$_{\rm{(1)}}$ & \textbf{73.8}$_{\rm{(1)}}$\\
       \citet{mager-etal-2020-gpt-too}-L&M'20 &33.0$_{\rm{(2)}}$& \textbf{37.7}$_{\rm{(1)}}$&63.9$_{\rm{(2)}}$ &  \textbf{94.0}$_{\rm{(1)}}$&\textbf{80.8} & 69.2 & 74.5$_{\rm{(2)}}$ & 69.8$_{\rm{(2)}}$ & 72.1$_{\rm{(2)}}$ & 73.5$_{\rm{(2)}}$ \\
       \citet{ijcai2020-542}& W'20 & \textbf{33.9}$_{\rm{(1)}}$ & 37.1$_{\rm{(3)}}$ & \textbf{65.8}$_{\rm{(1)}}$ &  93.7$_{\rm{(3)}}$ & 80.3 & \textbf{70.9} & \textbf{75.3}$_{\rm{(1)}}$ & 55.7$_{\rm{(3)}}$ & 64.0$_{\rm{(3)}}$ & 70.3$_{\rm{(3)}}$\\
       \bottomrule
    \end{tabular}}
    \caption{Main metric results.}
    \label{tab:mainres}
\end{table*}

\paragraph{Setup: data \&
%canonical 
metrics for system ranking} We obtain  test predictions of several state-of-the-art \textit{AMR-to-text generation systems} on LDC2017T10, 
%which has served as 
the main benchmark for this task:
%testing grounds 
%used over the recent years: 
(i) densely connected graph convolutional networks \citep{guo-etal-2019-densely}; (ii) 
%the system of 
\citet{ribeiro-etal-2019-enhancing}'s system that uses a dual graph representation; 
%.  Then we have 
two concurrently published models (iii) based on graph transformers \citep{DBLP:conf/aaai/CaiL20, TACL1805} and (iv) a model based on graph transformers that uses reconstruction information \cite{ijcai2020-542} in a multi-task loss; 
%. And 
finally, we obtain predictions of two system variants of \citet{mager-etal-2020-gpt-too}
that fine-tune
%. This system fine-tunes 
LMs and encode linearized graphs using (v) a large and (vi) a medium-sized LM. We  true-case all sentences and parse them with GSII.

%. Here, we have (iv) predictions from a large model and (v) based on a medium-sized model.

To put the results of \MF into perspective, we display the scores of several metrics that 
%align with the sentence-matching setup 
have been
previously used for AMR-to-text: \textsc{Bleu}, \textsc{Meteor}, \textsc{Chrf++}. We also
%Additionally, we
calculate 
%the recent 
BERTscore \cite{bert-score} with RoBERTa-large \cite{roberta}.\footnote{BERTscore computes an F1-score over a \textit{cosim}-based alignment of the contextual embeddings of paired sentences.}
%from the generated sentence and the reference sentence.}
%The 
Results are displayed in Table \ref{tab:mainres}, col.\ 3-6. \MF scores (col.\ 7-12) are divided into $Meaning$ (\resmatch using GSII) and $Form$ scores (based on GPT-2), and composite \MF scores with $\beta=1$ (harmonic) and $\beta=0.5$ (double weight on $\mathcal{M}$). 

%\paragraph{\resmatch upper-bound approximation} %Resmatch has an upper-bound. This is because, naturally, the parsers are not perfect and produce errors, and we can only approximate the human $f$ with $parse$. Therefore, 

As an upper-bound approximation
%oracle 
for \resmatch we propose parsing a gold sentence $s$ and comparing the result against the gold AMR $m$: $apprUB$ = metric(parse($s$),$m$).\footnote{%Essentially, t
This is the 
%same 
score of canonical parser evaluation. I.e.,
%This means that 
we would not expect the reconstruction 
%parse 
$m'$ of $s'$
to score higher than had we applied $parse$ to the original sentence: $metric(m',m)$ $\leq$ $metric(parse(s), m)$
= $apprUB$. %, where $s', s$ are the generated and original sentence, $m'=parse(s')$ is the reconstructed AMR $m'$, and $m$ the original AMR. 
This is an idealization, as we can imagine cases where the original sentence $s$ is more complex and thus more difficult to parse to an AMR than a simpler generated  paraphrase $s'$. Since we are interested in a very rough upper bound estimation, we abstract from such cases in our present work.}

\subsection{
%Enhanced 
Interpretability of system rankings} 
%with \MF}
\label{subsec:mainresults} 
\paragraph{Surface matching metrics lack differentiation and interpretability} 
%By inspecting 
Table \ref{tab:mainres} shows that
%we see that 
the baseline metrics tend to agree with each other on the ranking of systems, but there are also
%also exist 
differences, for example, BERTscore and \textsc{Meteor} select M'20 as the best performing system while \textsc{Bleu} and \textsc{ChrF++} select W'20. While certain differences may be due to individual metric properties, %of metrics as such, 
e.g., \textsc{Meteor} allowing inexact word matching of synonyms, 
%in general, 
the underlying factors
are difficult to assess, since the score differences between systems with switched ranks are 
%rather 
small, and none of these metrics can %hardly 
provide us with a meaningful interpretation of their score that would extend beyond shallow surface statistics. Hence,
%erefore, 
these metrics cannot give us much intuition about why and when one system may be preferable over another.

\paragraph{$Meaning$ vs.\ $Form$: How \MF explains system performance} We have seen
 that 
current
%the canonical 
metrics cannot provide us with convincing explanations as to why, e.g., W'20 should be preferred
%selected 
over M'20 (\textsc{Bleu}), or M'20 
%should be selected 
over W'20 (BERTscore). \MF score, however, tells a %much clearer 
story about how these systems differ, highlighting their complementary strengths by  
%We find that this issue appears %rectified by our \MF score. By 
disentangling $Meaning$ and $Form$ \citep{bender-koller-2020-climbing}: 
 W'20 displays the highest \resmatch score, i.e., AMRs constructed from its generations  
%\MF answers this question as follows:
%if a user desires a system that 
recover a maximum
%as much 
of the meaning contained 
in the input AMR.
%as possible. 
M'20, by contrast, outperforms all systems in $Form$ score. Looking at $\mathcal{M}\mathcal{F}_{1}$, the harmonic mean of both, both systems still occupy leading ranks, but W'20 falls back to 3rd rank, due to its weaker $Form$ score. 

Hence, given our metric principles, a user who cares about faithfulness to meaning, but less 
%but is not so 
%concerned with 
%proper 
about fluency,
%grammaticality, they 
should select W'20 (with higher \resmatch compared to M'20 by $\Delta$=1 point) -- 
%according to $\mathcal{M}\mathcal{F}_{0} =$ \resmatch: $\Delta$=1 point). On the other hand, if 
a user who desires a system that preserves meaning well but also 
%recovers 
%a good proportion of 
%the meaning contained in the input AMR well, but also
produces sentences of 
%generations of grammatically 
%significantly improved 
decent form,
%they should 
should select M(b)'20 
%(according to both 
(with $\mathcal{M}\mathcal{F}_{0.5}$ and $\mathcal{M}\mathcal{F}_{1}$ score differences against W'20 of $\Delta$=3.5 points and $\Delta$=8 points). Overall, \MF mostly agrees with BERTscore in the rankings of the teams. However, \MF's larger score differences between the systems, due to $Form$, are striking, prompting us to investigate the $Form$ predictions in closer detail (\S \ref{subsec:form}). %However, we 
We will see that using a different $Form$ predictor as well as a manual native speaker annotation clearly support our assessment of $Form$.

\subsection{On the quest for deep\underline{er} explanation and interpretation} \label{subsec:finegrained}

\begin{figure}[h]
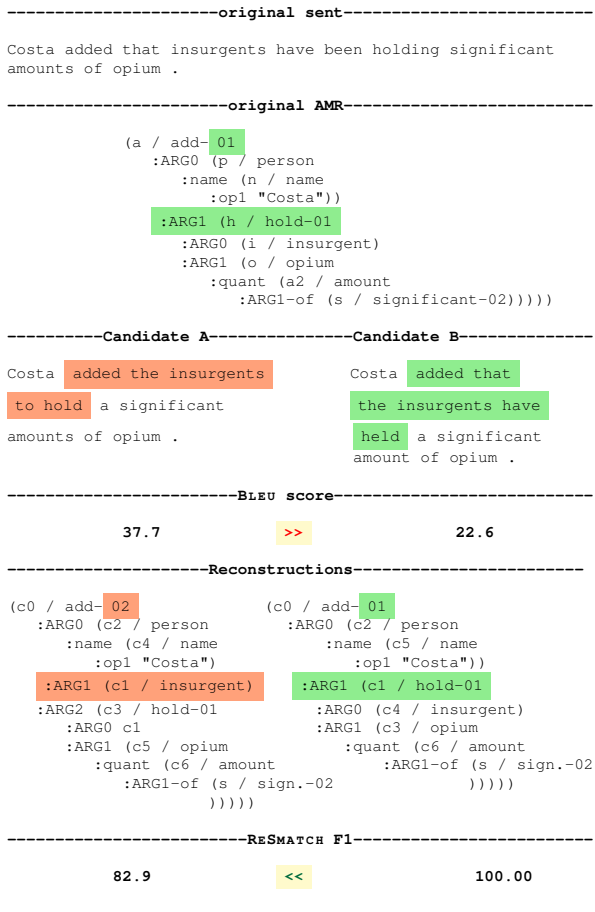


\begin{tiny}
\begin{Verbatim}[commandchars=\\\{\},codes={\catcode`$=3\catcode`^=7}]
\textbf{----------------------original sent--------------------------}

Costa added that insurgents have been holding significant 
amounts of opium .

\textbf{-----------------------original AMR--------------------------}

            (a / add-\colorbox{lightgreen}{01}
               :ARG0 (p / person 
                  :name (n / name 
                     :op1 "Costa"))
               \colorbox{lightgreen}{:ARG1 (h / hold-01}
                  :ARG0 (i / insurgent)
                  :ARG1 (o / opium
                     :quant (a2 / amount
                        :ARG1-of (s / significant-02)))))
                        
\textbf{----------Candidate A---------------Candidate B--------------}

Costa \colorbox{lightsalmon}{added the insurgents}        Costa \colorbox{lightgreen}{added that}
\colorbox{lightsalmon}{to hold} a significant             \colorbox{lightgreen}{the insurgents have}
amounts of opium .                  \colorbox{lightgreen}{held} a significant 
                                    amount of opium . 

\textbf{------------------------\textsc{Bleu} score---------------------------}

\textbf{            37.7            \colorbox{lemonchiffon}{\textcolor{red}{>>}}               22.6}

\textbf{---------------------Reconstructions------------------------}

(c0 / add-\colorbox{lightsalmon}{02}             (c0 / add-\colorbox{lightgreen}{01}
   :ARG0 (c2 / person        :ARG0 (c2 / person
      :name (c4 / name           :name (c5 / name
         :op1 "Costa")              :op1 "Costa"))
   \colorbox{lightsalmon}{:ARG1 (c1 / insurgent)}   \colorbox{lightgreen}{:ARG1 (c1 / hold-01}              
   :ARG2 (c3 / hold-01          :ARG0 (c4 / insurgent)
      :ARG0 c1                  :ARG1 (c3 / opium
      :ARG1 (c5 / opium            :quant (c6 / amount
         :quant (c6 / amount           :ARG1-of (s / sign.-02     
            :ARG1-of (s / sign.-02              )))))     
                     )))))
                     
\textbf{-------------------------\textsc{\resmatch} F1-------------------------}

\textbf{           82.9             \colorbox{lemonchiffon}{\textcolor{cadmiumgreen}{<<}}                 100.00}

\textbf{-------------------------------------------------------------}
   \end{Verbatim}
   \end{tiny}
\caption{Explainable $Meaning$ score (re-)ranking.}
\label{fig:costa}
\end{figure}

\resmatch can also provide us with \textbf{explanations for single-sentence (re-)rankings}. An example is shown in Figure \ref{fig:costa}. Here, the gold reference (both sentence and AMR) indicates that 
%there is 
a person named Costa 
%who 
\textit{adds} (as a communicative act\footnote{Sense add-01 w/ roles: Arg0: \textit{Speaker}; Arg1: \textit{Utterance}.}
%an act of speech) 
that some insurgents have been holding large amounts of opium. However, system generation A (which is higher ranked by \textsc{Bleu}) chooses a different sense of \textit{add}, add-02, which represents the action as an operation\footnote{Sense add-02 w/role set: Arg0: \textit{adder}; Arg1: \textit{thing being added}; Arg2: \textit{thing being added to}; Arg3: \textit{resulting sum}.}, which results in an incoherent or nonsensical meaning representation where the person Costa \textit{adds} (in the operational sense) the insurgent (as thing being added) to a circumstance to the effect that the insurgents hold a significant amount of opium. 
%triggers a very distinct interpretation, in the sense that the person named Costa \textit{added} (as an action that is not an act of speech) insurgents (possibly to a an already existing group of insurgents that is not explicitly mentioned) in order to be capable of holding more opium (\textit{he added the insurgents to hold...}). 
By contrast, system generation B preserves more of the gold AMR's meaning and clearly expresses that Costa performs an \textit{act of communication}
%speech} 
when he \textit{adds} something. 
 \resmatch ($\mathcal{M}\mathcal{F}_{\beta \rightarrow 0}$) is able to detect the meaning differences and assigns candidate B a significantly higher score than A, in fact, an S$^{2}$match score of 1.00.

\resmatch, when parameterized with fine-grained AMR evaluation metrics of  \citet{damonte-etal-2017-incremental},
%\MF 
can also
facilitate deeper insight into \textbf{how well system generations reflect or violate specific meaning aspects}. E.g., we can investigate
%as compared to the gold AMRs, e.g., 
a system's capacity to properly reflect negation (NEG); to generate correct
%appropriate 
surface forms for NEs (NER); assess how well a system captures
%represents 
coreference between entities (Coref); and whether or not the predicate-argument structures (SRL) of generated sentences appropriately reflect the source meaning. We apply these fine-grained AMR metrics to the \resmatch scores of systems displayed in Table \ref{tab:mainres} (see Appendix \ref{sec:appendix-damonte}), and observe, e.g., that 
R'19, which ranks last in the overall ranking, 
%which 
improves upon the best overall system by 3.4 points in NER recall and 1.9 points in F1. The analysis also corroborates that W'20 excels among competitors with best scores for coreference, SRL and negation, i.e., the more global aspects of sentence meaning.
Such information can be valuable for researchers for deeper system analysis and for practitioners aiming for specific use cases.

\section{Study II: Assess vulnerability of \MF}

\MF has two apparent vulnerabilities: first, it depends on a parser for reconstruction. We have used a SOTA parser that is on par with human IAA. Yet, we cannot exclude the possibility that it introduces unwanted errors in computing 
%the evaluation 
\MF scores.
%of \MF. 

Second, the $Form$ component is based on a LM and we have seen that it can change system rankings, even when it is discounted.\footnote{In Table \ref{tab:mainres}, both \MF with $\beta=0.5$ and $\beta=1.0$ slightly disagree with the ranks assigned by $Meaning$ only.} On the one hand, our LM was carefully selected, and other metrics such as
%(e.g., 
BERTscore also heavily depend on LMs. 
%Yet, on 
On the other hand, we cannot exclude the possibility that the changed rankings are unjustified.

Our next studies  %In this pilot study, we 
investigate these weak spots more closely. 
First, in \S \ref{subsec:parsequality},  we assess the outcome of \MF when using another parser and assess its potential portability to other text generation tasks by ablating the human gold graph and evaluate generated text against reference \textit{text}. In \S \ref{subsec:form} we 
%then
%Then we 
%discuss the result of 
conduct a human annotation study to assess whether the provided $Form$  rankings are justified.

\subsection{The parser: Achilles' heel of \MF?}\label{subsec:parsequality}

\paragraph{Using another parser} %\jo{[as recommended by rev 1 I add Groschwitz transition based parser in this experiment]} 
In this experiment %Now, we want to 
we assess the robustness of \resmatch against using  different parsers. This is important,
%an important point, 
since the metric and rankings could change with the parser.
%and/or users may have reasons to use different parsers for 
%the 
%reconstruction. 
Here, we would hope that the difference of using one competitive parser over another will not be too extreme, especially with regard to system rankings. To investigate this issue, we apply two alternative parsers: i) GPLA \cite{lyu-titov-2018-amr}, a neural graph-prediction system that jointly predicts latent alignments, concepts and relations, and ii) TTSA \cite{groschwitz-etal-2018-amr}, a neural transition-based parser that converts dependency trees to AMR graphs using a typed semantic algebra. We select GPLA and TTSA since they constitute technically quite
%quite 
distinct approaches compared to GSII.

\begin{table*}
\centering
    \scalebox{0.75}{\begin{tabular}{lrrrr|rrrr|rrrrr}
    \toprule
    &\multicolumn{4}{c}{\resmatch F1} & \multicolumn{4}{c}{ranks \resmatch} & \multicolumn{4}{c}{ranks $\mathcal{M}\mathcal{F}_{0.5}$}\\
     &TTSA & GPLA&GSII&GSII$^{\vardiamond}$ &  TTSA & GPLA&GSII&GSII$^{\vardiamond}$  & TTSA & GPLA&GSII&GSII$^{\vardiamond}$ \\
     $apprUB$ & 73.7 & 76.2  & 81.5 &  86.4 & 0 & 0 & 0 & 0& 0& 0 & 0 & 0\\ 
      R'19 & 66.9  & 70.1 & 71.9 & 72.3 & 7 & 7 & 6 & 6 & 5& 5 & 5 & 5\\
       G'19 & 69.7 & 72.2  & 73.9 & 73.7 & 3 & 3 & 3 & 4& 6&  6 & 6 & 6\\
Wb'20 & 67.3 &70.2 & 71.5 & 71.6 & 6 & 6 & 7  & 7&7 & 7 & 7 & 7 \\
       C'20 &69.1 & 70.4 & 72.2 & 73.4& 4 &5 & 5  & 5& 4 & 4 & 4& 4\\
       Mb'20 &68.9 & 70.5 & 73.7 & 74.2& 5&  4 & 4  &3 & 1 & 2 & 1 & 1\\
       M'20  & 69.8 & 72.5  & 74.5 & 75.1& 2 & 2 & 2   & 2& 2& 1 & 2 & 2 \\
       W'20   &70.5 & 73.1 & 75.3 & 75.4& 1 &1 &1 & 1 & 3&  3 & 3 & 3 \\
       \bottomrule
    \end{tabular}}
    \caption{Analysis of our metric using different parsers (GPLA, TTSA
    GSII) or ablating the gold parse by comparing the parsed generation against the parse (distant) source sentence (GSII$^\vardiamond$).}
    \label{tab:diffparser}
\end{table*}
The results are shown in Table \ref{tab:diffparser} (columns labelled GPLA, TTSA and GSII). 
%We see that
All variants tend to agree in the majority of their rankings\footnote{We observe one switch of ranks for TTSA-GPLA and GPLA-GSII and 2 rank switches for TTSA--GSII in \textsc{ReSmatch}, and no rank switch for TTSA-GSII and one switch for TTSA-GPLA and GPLA-GSII, for $\mathcal{MF}_{0.5}$.} (e.g., \resmatch$^{GPLA}$ vs.\ \resmatch$^{GSII}$ F1: Spearman's $\rho$ = 0.95, Pearson's $\rho$ = 0.96, p$<$0.001). When considering $\mathcal{M}\mathcal{F}_{\beta=0.5}$, the agreement further increases
%vulnerability further decreases 
(e.g., $\mathcal{M}\mathcal{F}_{0.5}^{GPLA}$ vs.\ $\mathcal{M}\mathcal{F}_{0.5}^{GSII}$: Spearman's $\rho$ = 0.95, Pearson's $\rho$ = 0.99, p$<$0.001).

However, while using TTSA or GPLA instead of GSII has little effect on the ranks,
%we see that 
%the 
the absolute scores can differ (e.g., W'20 70.5 F1 w/ TTSA, 73.1 F1 w/ GPLA and 75.3 F1 w/ GSII). Yet, we find
%conclude 
that none of the generation systems are unfairly treated by our main parser GSII since we observe (mostly uniform) increments from TTSA to GPLA and from GPLA to GSII. An unfair treatment could arise, e.g., if GSII  generates bad AMR reconstructions for specific NLG systems but not so for others. However, we do not observe such tendencies.
%major disagreements in rankings when using TTSA or GPLA parsers.

Hence we assume that
%Therefore, we may conjecture that 
GSII's score increments 
%may simply 
stem from the fact that GSII yields better reconstructions for all systems. In future work, we plan to explore 
%methods for 
parse quality control \cite{opitz-frank-2019-automatic,opitz-2020-amr} or ensemble parsing \cite{van-noord-bos-2017-meaning}, to gain more detailed information on the quality of the meaning reconstructions.
%is more benevolent to all generation systems.

\paragraph{Ablating the gold graph? Yes, we can.} 
%To %further 
In lack of a gold standard for the automatic reconstructions, we elicit some indirect answers and insight about the parser's quality, by considering the
%vet our parser, we pose the 
following question: \textit{What is the effect on system rankings when we replace the input gold graphs with \textit{automatic parses of the distant source sentence}?} If this effect is large, this will
%this would 
give us reasons to worry, as it would indicate that the parser is 
%possibly 
less reliable 
than expected given its 
%than we would have assumed by its 
high IAA with humans.
%on benchmark data. 
On the other hand, if we only see a minor effect,
%the effect is peripheral, 
this may
%could 
%further 
increase the trust in our parser and indicate that \textbf{\MF could be confidently applied for explainable evaluation in other generation tasks} (such as MT or summarization), where we do not have gold AMRs, and would have to parse both generated \textit{and} reference sentences.

The results of this experiment are displayed in Table \ref{tab:diffparser}: our standard setup is displayed in columns labeled GSII and the results of the setup where we replace the gold input graph with an automatic parse 
%ablate the gold graph 
is indicated by
%with 
GSII$^{\vardiamond}$. When 
%solely 
considering \resmatch scores, we see only one switched rank
between  Mb'20 and G'20 (3--4).
%: Mb'20 changes from 4 to 3 and G'20 from 3 to 4. 
However, note that
%When considering that 
the absolute F1 score $\Delta$ between these two systems is overall very small (GSII: 0.2; GSII$^{\vardiamond}$: 0.5). Overall, the scores do not tend to differ much when the gold graph is ablated, we observe rather small (mostly positive) changes in system scores (GSII $\rightarrow$ GSII$^{\vardiamond}$): 0.1 / 1.2 / 0.4 (min/max/avg).
In sum, we 
%may 
conclude from this experiment
that 
%putting more responsibility onto our parser by 
ablating the gold graph does not have a major effects on the scores and rankings. And when considering the $\mathcal{M}\mathcal{F}_{\beta=0.5}$ score, the ranking stays fully stable (the same holds true for $\mathcal{M}\mathcal{F}_{\beta=1}$. %\textcolor{red}{[I notice that we do not have MF with beta=1. How does it behave in comparison? And what is then "full MF score"?]}

\paragraph{Discussion}  We 
have shown that
%have seen that the 
metric rankings are fairly robust to using different parsers and that we do not necessarily depend on
%require a human generated
gold AMR graphs to compute the measure. This offers 
%points to potential 
prospects for \textbf{using \MF for an explainable assessment of systems that perform other kinds of text generation}. In order to measure $\mathcal{M}$, a parser
%Here, the parser can be 
could be applied to
%used to project 
both the generated and the reference text, to measure their agreement in the domain of
%in the domain of 
abstract meaning representation. 
This would in turn offer means for conducting
%allow to conduct a 
fine-grained meaning analysis of generation tasks where the reference is a natural language sentence (e.g., in MT). 

Note, however, that AMR, as of now, does not capture some facets of meaning that may be of interest in some generation tasks. For instance, it does not capture tense or aspect.  However, what we have investigated as a \textit{potential weakness} of \MF, namely the necessity to select a meaning parser, can also be viewed as a \textit{potential strength}. E.g., \citet{donatelli-etal-2018-annotation} show how tense and aspect can be captured with AMR. This indicates that \MF can indeed be used for a tense and aspect analysis of generated text -- if we parameterize it with a dedicated parser. Finally, if output and reference do not consist of single sentences, it may be apt to use a parser that constructs MRs for discourse (e.g., \textit{DRS} \cite{kamp1981theory}). 

In summary, we conclude that
%\jo{We can summarize that 
\MF, our proposed metric that aims to assess text generation quality by decomposing it into 
%, when viewed as a general principle to decompose the assessment of text generation quality into 
\textit{form} and \textit{meaning} aspects, is broadly applicable. However, different parser parametrizations may have to be considered in light of the specific nature of a generation task.

\subsection{The $Form$ component of \MF}\label{subsec:form}

In \S %Section
\ref{subsec:mainresults}, we have seen that the $Form$ %component 
aspect of \MF can change %the 
system ranks. Notably, it has 
promoted
%pushed \MF to select 
M'20 as the best generation system, outranking
%over 
W'20 (in agreement with BERTscore), whereas W'20 is selected by \textsc{Bleu} or \resmatch. Now, we aim to investigate
%want to see 
whether these impactful decisions of the $Form$ component were justified.

\paragraph{Human annotation} We ask a native speaker of English to 
%annotate
rate 50 paired 
%sentences 
generations of M'20 and W'20, 
%with respect to 
%their %structural 
%well-formedness, 
considering only grammaticality and fluency.\footnote{The annotator was explicitly instructed
%asked to 
not to consider 
whether a sentence `makes sense', by presenting the \textit{Green ideas sleep furiously} example as free from structural error.} We give 
more 
detail 
%on this annotations 
and provide examples in Appendix \ref{sec:appendix-annoation}. The annotator agreed in 42 of 50 pairs with the preference %as 
predicted by GPT-2 
%which is 
(a significant result: binomial test p$<$0.000001). We find that the M'20 and Mb'20 generations %indeed 
%appear 
are considerably better on the surface level, compared to generations of all other systems. For instance, the best system according to $Meaning$,
%on the meaning level, 
W'20, frequently produces inflection mishaps: \textit{Their hopes for entering the heat \underline{is} already in-sight}, while we find few %little of 
such violations with M'20 (here: \textit{Their hopes for entering the heat \underline{are} already in sight}). We also find errors with adverbials,
%errors to varying degrees, 
e.g., W'20 writes \textit{They are \underline{the most} indoor training at home}, while M'20 writes \textit{They are \underline{most} trained indoors at home.} Arguably both 
%of these 
sentences are not %of 
perfect 
%form 
%(correct: \textit{mostly}), 
but the second %sentence 
is 
substantially more well-formed.

\begin{table}
    \centering
    \scalebox{0.62}{
    \begin{tabular}{lrrrrrrr}
    \toprule
            & R'19 & G'20 & Wb'20 & C'20 & Mb'20 & M'20 & W'20 \\
            \midrule
         GPT-2 & 51.6$_{\rm{(4)}}$ & 47.1$_{\rm{(6)}}$ & 49.5$_{\rm{(5)}}$ & 51.9$_{\rm{(4)}}$ & 74.0$_{\rm{(1)}}$ &69.8$_{\rm{(2)}}$ & 55.7$_{\rm{(3)}}$ \\
         BERT & 43.4$_{\rm{(6)}}$ & 40.6$_{\rm{(7)}}$ & 50.4$_{\rm{(4)}}$ & 44.7$_{\rm{(5)}}$ &71.4$_{\rm{(1)}}$ & 71.0$_{\rm{(2)}}$ & 55.9$_{\rm{(3)}}$ \\
         \bottomrule
    \end{tabular}
    }
    \caption{$Form$ scores %of systems 
    when using a different LM.}
    \label{tab:difflm}
\end{table}
\paragraph{Using a different LM} The human study indicates that GPT-2 is accurate to 84\%
%mostly right 
when %it 
favoring one sentence over the other, with respect to fluency and grammaticality. However, when considering that there is a 
%recent
trend to building systems 
%that are 
based on fine-tuned LMs, we need to assess whether they may be favored (too) much if $Form$ is parameterized with a same or a highly similar LM to the one used by the NLG model.
%compared to the LM these systems use for tuning. 
We find such a case in M'20: 
while it was
%on one hand, they did 
not fine-tuned with the same GPT-2 that
%which 
we used for $Form$ assessment,
%prediction, 
they 
%but they 
fine-tuned their model with its siblings GPT-2-medium and GPT-large, which may share 
%great 
structural similarities. Therefore, we also use BERT for  $Form$ assessment.
%prediction. 
The results in Table \ref{tab:difflm} support the %unambiguous 
conclusion from the human annotation: by large margins, both M'20 and Mb'20 deliver generations that are of significantly improved form and both agree on the group of the three best %three 
systems. Note that this insight can be provided by $\mathcal{M}\mathcal{F}_{\infty}$, but it cannot be carved out by %using the 
conventional metrics, since these do not disentangle
%prohibit us from disentangling 
$Form$ and $Meaning$.

\section{Related work}\label{sec:relwork}

Traditionally, the performance of NLG systems has been evaluated with word n-gram matching metrics such as the popular \textsc{Bleu} metric in MT \cite{papineni-etal-2002-bleu} or Rouge \cite{lin-2004-rouge} in document summarization.  Yet, such metrics suffer from several well-known issues  \cite{novikova-etal-2017-need, nema-khapra-2018-towards,nemasai2020survey}. E.g., due to their symbolic matching strategy they cannot account for paraphrases. Recently, unsupervised \cite{bert-score} or learned metrics \cite{sellam2020bleurt,zhoucompare} based on contextual language models have been proposed. For example, 
%the 
BERTscore \cite{bert-score} %metric 
uses BERT \cite{devlin-etal-2019-bert} to encode candidate and reference and computes a
%computing the 
score based on a cross-sentence word-similarity alignment. Compared with \textsc{Bleu}, it is computationally more 
expensive but tends to show higher 
 %increase the 
 agreement with humans. 
 However, \textit{all} of the aforementioned metrics return 
 %a 
 scores that are hardly interpretable and we cannot tell what exactly they have measured. 
 
These problems carry over to the evaluation of AMR-to-text generation:
 %\jo{These problems also occur when evaluating AMR-to-text:
 \citet{may-priyadarshi-2017-semeval} find that \textsc{Bleu} does not well correspond to human ratings of generations from AMR, and \citet{manning-etal-2020-human} show through
 %in their 
 human analysis that none
 %all 
 of the 
 existing automatic metrics can provide nuanced views on generation quality. Our proposal \MF 
 %makes a steps 
 takes a first step to address these issues by aiming at a clear separation of form and meaning, as called for by \citet{bender-koller-2020-climbing}.

First attempts of assessing semantic generation quality have been examined in MT using semantic role labeling \cite{lo-2017-meant} or WSD and NLI \cite{carpuat-2013-semantic,poliak-etal-2018-evaluation}, in-between lies SPICE that evaluates caption generation via inferred semantic propositions \cite{10.1007/978-3-319-46454-1_24}. Just like \MF, SPICE relies on automatic parses (a dependency parse of the caption and a scene graph predicted for the image) to evaluate content overlap of image and caption. Thus, SPICE is a direct precursor of an NLG metric in V\&L that relies on automatically produced structured representations. 
%In this aspect, we add 
Our work extends this
%on the 
previous work by showing ways of probing potentially harmful effects of incorporating automatic parsing components.

\section{Conclusion}

We propose \MF score, a new %linguistically motivated 
metric for evaluation of text generation from (abstract) meaning representation. The metric is built on two pillars: $Form$ measures grammaticality and fluency of the produced sentences and $Meaning$ assesses to what extent the
%how much 
meaning of 
the input AMR is reflected in the produced sentence. We show
%saw 
that \MF has the potential to yield fine-grained %system 
performance assessment that go beyond what conventional metrics can provide. Using its $\beta$-parameter, %allows researchers to decompose 
\MF can be decomposed into complementary views -- $Meaning$ and $Form$ --
%in either of the two parts, 
paving the way for custom gauging and selection of NLG
%text generation 
systems. We have seen
%observe 
that \MF %score 
corresponds well to
%relates to
%could potentially be interpreted as 
BERTscore when rankings systems, but 
%mitigates its opaqueness by offering to focus on meaning aspects disentangled from form properties.
overcomes its opaqueness by disentangling $Meaning$- and $Form$-related quality aspects.
%Conversely, and 
In sharp contrast to BERTscore, the $Form$ component of \MF dispenses with
%functions without 
%that does not 
%relying on
%resorting to 
string matching 
%of generated sentences 
against %their 
reference sentences, offering an assessment independent of lexical alignment.

An important
%A critical 
hyperparameter of our metric is 
the required AMR parsing component
%the dependency on a parser
for meaning reconstruction. 
%To alleviate this issue, we used the latest state-of-the-art parser in our experiments. 
%In addition, %Additionally, 
We investigate 
the impact of its choice
%this dependency 
by choosing alternative high-performing
%trying out 
%a different
parsers. Our study shows that absolute metric scores tend to increment when using a better parser, 
%are used, 
while
%and that %but the 
system rankings are
%of systems 
%stay 
quite stable. Furthermore, we outline the potential of \MF to  
extend to further
%be used in other 
text generation tasks, by ablating the human gold graph from the evaluation, such that the metric score can be computed from candidate and reference text alone.
Since benchmarking of systems needs deeper exploration, we  recommend
%consider 
%the usage of 
\MF score %as a promising score 
to obtain better diagnostics and explainability of text generation systems, including, but not limited to (A)MR-to-text.
%generated texts
%another interesting 
%as a promising use case.
\section*{Acknowledgments}

We are grateful to three anonymous reviewers for their valuable comments that have helped to improve this paper. This work has been supported by the \textit{Deutsche Forschungsgemeinschaft (DFG)} through the project \textit{ExpLAIN}, FR 1707/4-1 as part of the RATIO Priority Program (SPP-1999).
\bibliography{anthology,eacl2021}
\bibliographystyle{acl_natbib}

\clearpage
\appendix
\section{Appendices}
\label{sec:appendix}
\subsection{On the soundness of comparing generated sentences in the AMR domain}
\label{sec:appendix-proof}

\begin{figure}
    \centering
    \includegraphics[width=0.9\linewidth]{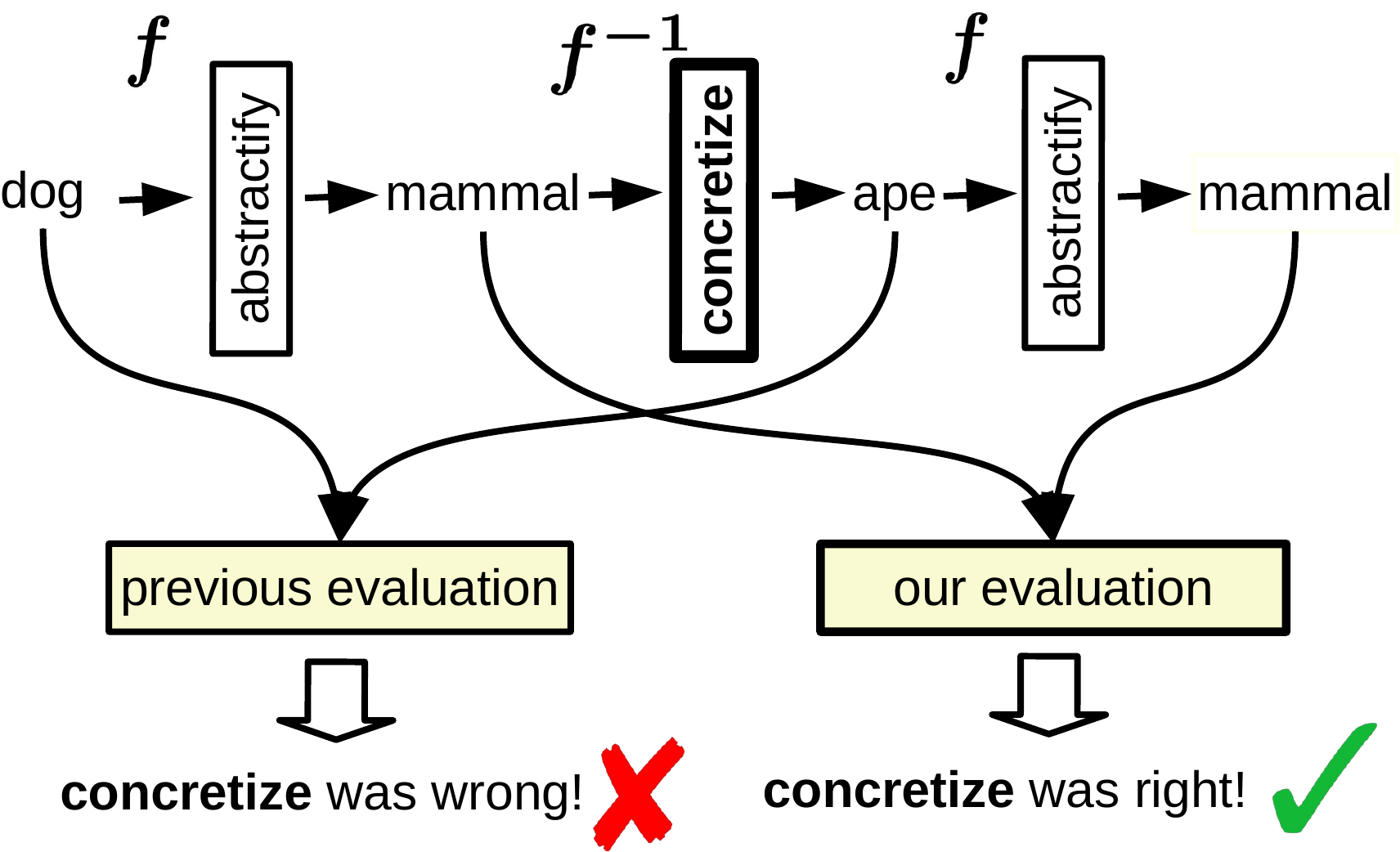}
    \caption{A critical issue and its alleviation.% may arise in the scenario where one assesses the output of a function that maps from rather abstract objects to more concrete objects.%, when evaluating the performance of a function $f^{-1}$ by comparing $x'=f^{-1}(f(x))$ against $x$, where $f$ is an surjective function. We alleviate the issue by comparing $f(x')= f(f^{-1}(f(x)))$ against $f(x)$.
    }
    \label{fig:problem2}
\end{figure}
First, we provide a simple example for our argument (it is safer to compare texts generated from AMR in the AMR domain) and then a simple proposition together with its proof. The example is displayed in Fig.\ \ref{fig:problem2}, where, 
%in an analogue 
similar to AMR-to-text, we see a (surjective) function that generates concrete objects from abstract objects (e.g., $mammal \rightarrow \{dog, mouse, cow\}$). Now, 
imagine
%consider that 
we are given $mammal$ and are tasked with generating a single concrete instance. How can we assess whether our output is correct? 
%We observe that 
We cannot safely assess this
%safely happen 
by testing whether the output (e.g., $cow$) is the same as another instance of $mammal$ (e.g., $dog$). Instead, we can 
%use $f^{-1}$ as a right-inverse function, 
re-apply the abstraction $f$ to \textit{cow} and conduct the comparison safely in the abstract domain. 

%Therefore, it is better to use $f^{-1}$ as a right-inverse function of $f$.  To see this, 
\begin{prop}
a) The canoncical AMR-to-text evaluation setup, that matches generated sentence $s'$ to distant source sentence $s$, is not well defined. 
b) This issue can be alleviated by grounding the evaluation in the AMR domain by re-appling parse, abstaining from direct use of $s$ (thereby using AMR-to-text generation as a right inverse function).
\end{prop}
\textbf{Proof.} Let $X$ be a set of concrete objects (e.g., sentences) and $f$ a (surjective) function from $X$ to $Y$ (e.g., `sent-to-AMR'), where $Y$ contains abstract objects (e.g., AMRs), s.t.\ $|Y| < |X|$. Then, using $f^{-1}: Y \rightarrow X$ (e.g., `AMR-to-sent')) as right-inverse is well-defined: $f \circ f^{-1}  = id_Y$ (Proposition b), but using it solely as left-inverse (as done in previous evaluation) does not guarantee a well-defined result: $f^{-1} \circ f \neq id_X$ (Proposition a). $\qed$

\subsection{Form predictor selection experiment}\label{sec:appendix-webnlgexp}
To estimate how well they are able to assess $Form$, we make use of human-assigned scores for data from the WebNLG task as provided by
%As data we use the the data by 
\citet{gardent-etal-2017-webnlg}.
It
%that 
contains grammaticality and fluency judgments by humans for more than 2000 machine-generated sentences. We report the F1 score, both for grammaticality and fluency, by converting the human assessment scores to %of the human 
$accept$ predictions, and using them as a gold standard to evaluate
%and 
the LM-based $accept$ predictions over (i) all 12k sentence pairs\footnote{This includes
%consider 
all generated sentences from a given input, as provided by \citet{gardent-etal-2017-webnlg,shimorina2017webnlg}} and (ii) only the 5k sentence pairs where both grammaticality and fluency
%parts 
where either rated as `perfect' (max.\ score) or `poor' (min.\ score) by the human.\footnote{The ratings are based on a 3-point Likert scale.}

\begin{table}
    \centering
    \scalebox{0.7}{
    \begin{tabular}{l|rr|rr}
    \toprule
        & \multicolumn{4}{c}{F1 score }  \\   
        & \multicolumn{2}{c}{grammaticality } & \multicolumn{2}{c}{fluency}\\
        \midrule
        LM & poor/perfect & all & poor/perfect & all \\
        \midrule
        GPT2 & \textbf{0.80}&\textbf{0.74} & \textbf{0.80} & 0.71 \\
        GPT2-distill & 0.79 & 0.73 &0.76&0.70 \\
        BERT & \textbf{0.80} & 0.72 & \textbf{0.80}& \textbf{0.72}\\
        RoBERTa & 0.66 & 0.72 & 0.69 & \textbf{0.72}\\
        \bottomrule
    \end{tabular}}
    \caption{Results for assessing the $Form$ score  prediction (corpus-level) of different LMs for NLG-generated sentences against  humans judgements (separated by grammaticality and fluency); all: all 12k generated sentences vs. 'poor/perfect': the 5k instances of best/worst generations in both grammaticality and fluency.}
    \label{tab:predselect}
\end{table}

The results are displayed in Table \ref{tab:predselect} and show (i) that the LMs lie very close to each other with respect to their capacity to predict fluency and grammatically, and (ii) that both fluency and grammaticality can be predicted fairly well.

\subsection{\resmatch with fine-grained meaning metrics} \label{sec:appendix-damonte}
Using \citet{damonte-etal-2017-incremental}'s metric suite for fine grained semantic system analysis, we obtain fine-grained results with respect to various meaning aspects of system performance. The results are shown in Table \ref{tab:finegrained}.

\begin{table}
\centering

  \scalebox{0.52}{ 
    \begin{tabular}{l|rrrrrrrrrrrr}
    \toprule
    &\multicolumn{3}{c}{Reentrancies}&\multicolumn{3}{c}{SRL} & \multicolumn{3}{c}{negation} & \multicolumn{3}{c}{NER}\\
    \cmidrule{2-13}
     &P&R&F1&P&R&F1&P&R&F1&P&R&F1\\
     \midrule
     $apprUB$ & 72.1 & 60.7 & 65.9 & 77.7 & 73.5 & 75.5  &  88.6 &  70.5 &  78.5  & 82.2 & 80.1 & 81.1\\ \midrule
      R'19  & 63.7 & 50.3 & 56.2 & 71.1 & 62.4 & 66.4 &  72.1 & 50.6& 59.5
  &82.2 & \textbf{70.7} &\textbf{ 76.0} \\
       G'19 &  66.9 &52.9 & 59.1 & 73.7& 64.9& 69.0 & 75.0& 51.5& 61.1 &  78.6 & 68.9& 73.5\\
       Wb'20 &  67.6 & 51.5 & 58.4 & 75.1 & 63.6 & 68.9 & 74.3 & 49.7 & 59.6 & 86.5 & 60.3 & 71.0\\
       C'20 &   66.1 & 52.4 & 58.4
 &73.4  & 64.8  &  68.8 & 78.3 & 54.2 & 64.1 &  80.8 & 67.2& 73.4\\
       Mb'20 & 65.9 & 53.2 & 58.9 & 74.3 & 65.7 & 69.8 & 70.6 & 45.5 & 55.3 
 & 82.6 & 69.4 & 75.4 \\
       M'20 & 67.9 & 53.3 & 59.7 & \textbf{76.4} & 66.5 & 71.1 & 73.7 & 53.9 & 62.3 & \textbf{82.8} & 68.3 & 74.9 \\
W'20 & \textbf{68.8 }& \textbf{55.7} & \textbf{61.6} &  76.1 &  \textbf{68.1} & \textbf{71.9} & \textbf{79.2} & \textbf{55.1 }& \textbf{65.0} &  82.4 & 67.3 & 74.1\\
\bottomrule
    \end{tabular}}
    
    \caption{Fine-grained results using $\mathcal{M}\mathcal{F}_{0}$ parameterized with metrics proposed by \citet{damonte-etal-2017-incremental}. 
    }
    \label{tab:finegrained}
\end{table}
In sum, the system of W'20 appears to be the clear winner in most aspects of meaning. This is intuitive, since the system has been trained with an auxiliary signal that provides information on how well an AMR can be reconstructed from the generated sentence.

\subsection{\resmatch explains negation error}\label{sec:appendix-negationerror}

\begin{figure}
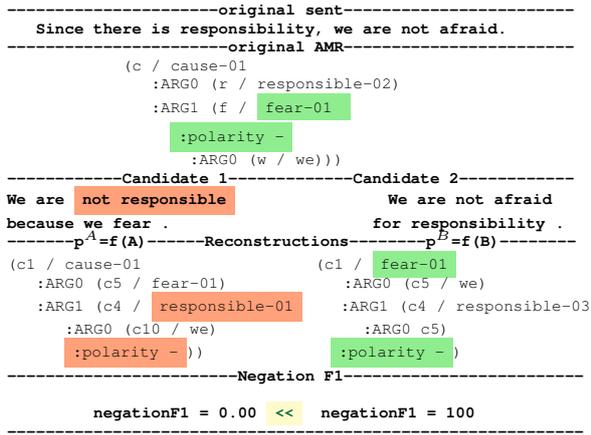

\begin{tiny}
\begin{Verbatim}[commandchars=\\\{\},codes={\catcode`$=3\catcode`^=7}]
\textbf{----------------------original sent------------------------}
   \textbf{Since there is responsibility, we are not afraid.}
\textbf{-----------------------original AMR------------------------}
            (c / cause-01
               :ARG0 (r / responsible-02)
               :ARG1 (f / \colorbox{lightgreen}{fear-01 }
                 \colorbox{lightgreen}{:polarity -}
                   :ARG0 (w / we)))
\textbf{------------Candidate 1-------------Candidate 2------------}
\textbf{We are \colorbox{lightsalmon}{not responsible}                We are not afraid}
\textbf{because we fear .                     for responsibility .}
\textbf{-------p$^A$=f(A)------Reconstructions--------p$^B$=f(B)--------}
(c1 / cause-01                  (c1 / \colorbox{lightgreen}{fear-01}
   :ARG0 (c5 / fear-01)            :ARG0 (c5 / we)        
   :ARG1 (c4 / \colorbox{lightsalmon}{responsible-01}    :ARG1 (c4 / responsible-03
      :ARG0 (c10 / we)               :ARG0 c5)
      \colorbox{lightsalmon}{:polarity -}))             \colorbox{lightgreen}{:polarity -})
\textbf{------------------------Negation F1-------------------------}

         \textbf{negationF1 = 0.00 \colorbox{lemonchiffon}{\textcolor{cadmiumgreen}{<<}}  negationF1 = 100}
\textbf{------------------------------------------------------------}
   \end{Verbatim}
   \end{tiny}
\caption{Explained negation confusion.}
\label{fig:negationconfusion}
\end{figure}

In Figure \ref{fig:negationconfusion}, both systems struggle to fully capture the meaning of the original AMR $f(s)$. However, the system based on GPT medium (Mb'20) erroneously assesses that \textit{we are not responsible} and \textit{we fear}. However, quite the opposite is true: the gold graph and gold sentence states that \textit{there is responsibility} and \textit{there is no fear}. This important facet of meaning is better captured by C'20. The reconstruction shows that it reflects the gold negated concepts much better and does not distort facts that are core to the meaning. In consequence, the negation F1 is zero for the left sentence with the distorted facts and maximum for the sentence that sticks true to the facts.

\subsection{\resmatch explains SRL error}\label{sec:appendix-srlerror}
\begin{figure}[ht]
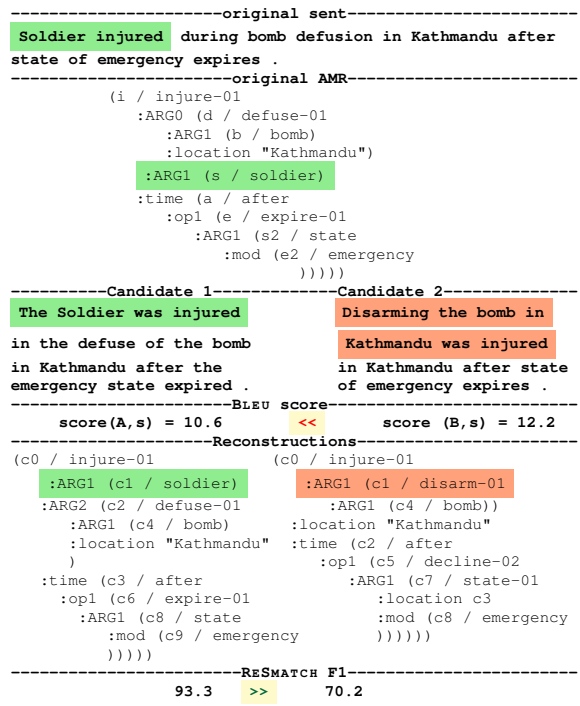


\begin{tiny}
\begin{Verbatim}[commandchars=\\\{\},codes={\catcode`$=3\catcode`^=7}]
\textbf{----------------------original sent------------------------}
\textbf{\colorbox{lightgreen}{Soldier injured} during bomb defusion in Kathmandu after }
\textbf{state of emergency expires .}
\textbf{-----------------------original AMR------------------------}
          (i / injure-01
             :ARG0 (d / defuse-01
                :ARG1 (b / bomb)
                :location "Kathmandu")
             \colorbox{lightgreen}{:ARG1 (s / soldier)}
             :time (a / after
                :op1 (e / expire-01
                   :ARG1 (s2 / state
                      :mod (e2 / emergency
                              )))))
\textbf{----------Candidate 1-------------Candidate 2--------------}
\textbf{\colorbox{lightgreen}{The Soldier was injured}         \colorbox{lightsalmon}{Disarming the bomb in}}
\textbf{in the defuse of the bomb         \colorbox{lightsalmon}{Kathmandu was injured} }
\textbf{in Kathmandu after the            in Kathmandu after state}
\textbf{emergency state expired .         of emergency expires .}
\textbf{-----------------------\textsc{Bleu} score--------------------------}
\textbf{     score(A,s) = 10.6       \colorbox{lemonchiffon}{\textcolor{red}{<<}}      score (B,s) = 12.2}
\textbf{---------------------Reconstructions-----------------------}
(c0 / injure-01            (c0 / injure-01
   \colorbox{lightgreen}{:ARG1 (c1 / soldier)}     \colorbox{lightsalmon}{:ARG1 (c1 / disarm-01}
   :ARG2 (c2 / defuse-01         :ARG1 (c4 / bomb))
      :ARG1 (c4 / bomb)      :location "Kathmandu" 
      :location "Kathmandu"  :time (c2 / after
      )                         :op1 (c5 / decline-02                 
   :time (c3 / after               :ARG1 (c7 / state-01
     :op1 (c6 / expire-01             :location c3
       :ARG1 (c8 / state              :mod (c8 / emergency
          :mod (c9 / emergency        ))))))      
          )))))        
\textbf{------------------------\textsc{\resmatch} F1------------------------}
            \textbf{     93.3   \colorbox{lemonchiffon}{\textcolor{cadmiumgreen}{>>}}     70.2}
\textbf{-----------------------------------------------------------}
   \end{Verbatim}
   \end{tiny}
\caption{Explained SRL confusion.}
\label{fig:ranking1srl}
%\vspace{-10mm}
\end{figure}

Figure \ref{fig:ranking1srl} shows an example, were \resmatch ranks two generated candidate sentences differently compared to \textsc{Bleu}. In this case, gold sentence and gold AMR both express that there is some soldier who tried to defuse a bomb and got injured in the process. Clearly, candidate generation A captures the meaning better, in fact, it captures it almost perfectly. However, since the surface text deviates from the gold sentence, \textsc{Bleu} overly penalizes this generation and assigns a very low score of 10.6 points. In contrast, candidate B matches the surface slightly better (12.2 points), but distorts the meaning: it does not contain any information about the soldier and states that \textit{Disarming was injured}, which is grammatically correct, but semantically wrong, or even non-sense. 

We see that the surface matching metric cannot explain its scores (beyond superficial statistics) and delivers a ranking that does not appropriately reflect the performance of the generation systems. However, \resmatch shows that the gold parse and the parse of candidate A agree with each other in the central \textit{ARG1}-role of the main predicate  \textit{injure-01}: \textit{it is the soldier who got injured}. On the other hand, in the reconstruction of the AMR of candidate B, the \textit{ARG1} argument is filled differently: \textit{it is the disarmament that gets injured}.

This assessment allows \resmatch to increment the score for generation A by a large margin, from 10.6 (\textsc{Bleu}) to 93.3 points (\resmatch), expressing substantial agreement in meaning with the gold. The score for the candidate generation B also gets incremented -- but it gets incremented much less, only to 70.2 points, expressing good to mediocre agreement. Thus, by detecting the SRL confusion, \resmatch re-ranks the candidate generation such that the resulting ranking is more appropriate. 

\subsection{Annotation study for form assessment}\label{sec:appendix-annoation}

\begin{figure*}[t!]
    \begin{tiny}
    \centering
    \begin{Verbatim}
Sys (W'20): He also said that our athletes do n't very use of competition under strong sunlight .
Corr (human): He also said that our athletes are not very used to competition under strong sunlight .
---->  not acceptable
        
Sys (W'20): Sheng Chen , the 6 th position of Hubei province , who was totally scored 342.60 at 342.60 points this year , 
is a temporary position .
Corr (human): Sheng Chen , the 6 th position of Hubei province , who has totally scored 342.60 points this year , 
is in a temporary position .
---->  not acceptable
        
Sys (W'20): The Chinese competitors are Lan Wei and Sheng Chen , qualify semi - final .
Corr (human):  The Chinese competitor Lan Wei and Sheng Chen qualify for the semi - final .
---->  acceptable
        
Sys (M'20): Fengzhu Xu won many championships in international competition before .
Corr (human): Fengzhu Xu won many championships in international competitions before .
---->  acceptable
    \end{Verbatim}
    \end{tiny}
    
    \caption{Sentences of  flawed form.  \texttt{--->} refers to the binary acceptability judgment (Eq. \ref{eq:tol}).}
    \label{fig:badform}
\end{figure*}

\paragraph{Annotator and annotation} The English native speaker (UK) annotated 50 paired sentences of M'20 and W'20. They were presented in shuffled order and the annotator was tasked with assigning a nominal number, starting from zero, that indicates the amount of grammatical or fluency issues as assessed by the native speaker. Additionally, the human was asked to provide a correction.

\paragraph{Examples of sentences of flawed form.} See Figure \ref{fig:badform}.

\end{document}